\documentclass[lettersize,journal]{IEEEtran}
\usepackage{amsmath,amsfonts}
\usepackage{algorithmic}
\usepackage{algorithm}
\usepackage{array}
\usepackage[caption=false,font=normalsize,labelfont=sf,textfont=sf]{subfig}
\usepackage{textcomp}
\usepackage{stfloats}
\usepackage{url}
\usepackage{verbatim}
\usepackage{graphicx}
\usepackage{cite}
\usepackage{bm}
\usepackage{booktabs}
\usepackage{multirow}
\usepackage{color}
\usepackage[cmyk]{xcolor}
\begin{document}

\title{Fully Spiking Actor Network with Intra-layer Connections for Reinforcement Learning}

\author{Ding Chen, Peixi Peng, Tiejun Huang,~\IEEEmembership{Senior Member,~IEEE,} and Yonghong Tian,~\IEEEmembership{Fellow,~IEEE}
\thanks{This paper has received final approval for publication as a [Regular Paper] in the IEEE Transactions on Neural Networks and Learning Systems.}
\thanks{Ding Chen is with the Department of Computer Science and Engineering, Shanghai Jiao Tong University, Shanghai 200240, China (email: lucifer1997@sjtu.edu.cn).}
\thanks{Peixi Peng, Tiejun Huang, and Yonghong Tian are with the Department of Computer Science and Technology, Peking University, Beijing 100871, China (email: pxpeng@pku.edu.cn; tjhuang@pku.edu.cn; yhtian@pku.edu.cn).}
\thanks{Ding Chen, Peixi Peng, and Yonghong Tian are also with Network Intelligence Research,
PengCheng Laboratory, Shenzhen 518066, China.}
\thanks{Yonghong Tian is also with the School of Electronics Computer Engineering, Peking University, Shenzhen 518055, China.}}

\maketitle

\begin{abstract}
With the help of special neuromorphic hardware, spiking neural networks (SNNs) are expected to realize artificial intelligence (AI) with less energy consumption. It provides a promising energy-efficient way for realistic control tasks by combining SNNs with deep reinforcement learning (DRL). In this paper, we focus on the task where the agent needs to learn multi-dimensional deterministic policies to control, which is very common in real scenarios. Recently, the surrogate gradient method has been utilized for training multi-layer SNNs, which allows SNNs to achieve comparable performance with the corresponding deep networks in this task. Most existing spike-based RL methods take the firing rate as the output of SNNs, and convert it to represent continuous action space (i.e., the deterministic policy) through a fully-connected (FC) layer. However, the decimal characteristic of the firing rate brings the floating-point matrix operations to the FC layer, making the whole SNN unable to deploy on the neuromorphic hardware directly. To develop a fully spiking actor network without any floating-point matrix operations, we draw inspiration from the non-spiking interneurons found in insects and employ the membrane voltage of the non-spiking neurons to represent the action. Before the non-spiking neurons, multiple population neurons are introduced to decode different dimensions of actions. Since each population is used to decode a dimension of action, we argue that the neurons in each population should be connected in time domain and space domain. Hence, the intra-layer connections are used in output populations to enhance the representation capacity. This mechanism exists extensively in animals and has been demonstrated effectively. Finally, we propose a fully spiking actor network with intra-layer connections (ILC-SAN). Extensive experimental results demonstrate that the proposed method outperforms the state-of-the-art performance on continuous control tasks from OpenAI gym. Moreover, we estimate the theoretical energy consumption when deploying ILC-SAN on neuromorphic chips to illustrate its high energy efficiency.
\end{abstract}

\begin{IEEEkeywords}
Spiking neural networks, reinforcement learning, brain-inspired intelligence, neuromorphic engineering, non-spiking neurons, intra-layer connections.
\end{IEEEkeywords}

\begin{figure}[!t]
\centering
\includegraphics[width=\columnwidth]{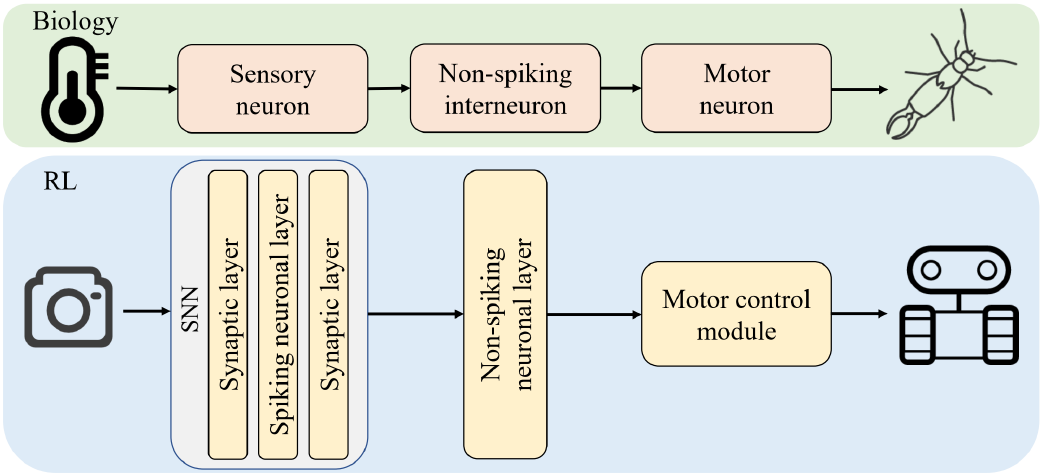}
\caption{The correspondence diagram between our method and the sensory motor neuron pathway.}
\label{fig:sensorimotor}
\end{figure}

\section{Introduction}
\IEEEPARstart{R}{ecently}, guided by the brain, neuromorphic computing has emerged as one of the most promising types of computing architecture, which could realize energy-efficient AI through spike-driven communication~\cite{kuzum2013synaptic,indiveri2015memory,roy2019towards}. The research efforts of neuromorphic computing not only facilitate the emergence of large-scale neuromorphic chips~\cite{merolla2014million,davies2018loihi,furber2014spinnaker}, but also promote the development of SNNs~\cite{maass1997networks,fang2021incorporating,fang2021deep}. In this context, the field of neuromorphic computing is a close cooperation organic whole between hardware and algorithm.

\IEEEpubidadjcol

An accumulating body of research studies shows that SNNs can be used as energy-efficient solutions for robot control tasks with limited on-board energy resources~\cite{mahadevuni2017navigating,bing2018end,bing2018survey}. To overcome the limitations of SNNs in solving high-dimensional control problems, it would be natural to combine the energy-efficiency of SNNs with the optimality of deep reinforcement learning (DRL), which has been proven effective in extensive control tasks~\cite{mnih2015human,silver2018general}. Since rewards are regarded as the training guidance in reinforcement learning (RL), several works~\cite{fremaux2013reinforcement,fremaux2016neuromodulated} employ reward-based learning using three-factor learning rules. However, these methods only apply to shallow SNNs and low-dimensional control tasks, or require manual tuning of the network architecture and numerous hyper-parameters related to neurons and synapses for each scenario~\cite{bellec2020solution}. Besides, the surrogate gradient method~\cite{lee2016training} provides a promising way to train deep SNNs. In a typical control task, the agent needs to learn multi-dimensional deterministic policies. In the deterministic policy learning, the RL methods~\cite{lillicrap2015continuous,fujimoto2018addressing,haarnoja2018soft} often utilize the actor-critic framework, where the actor network maps the state to the continuous action space and the critic network is used to represent state-action value (i.e., Q-value). Note that the critic network is just used for training, and only the actor network is deployed on the hardware of the application, hence it is necessary to develop a spiking actor network (SAN) as the energy-efficient solution for control tasks. Such network structure includes a deep critic network and a SAN, also known as the hybrid framework. However, the existing methods based on hybrid framework~\cite{tang2021deep,zhang2022multiscale} take the firing rate\footnote{ It should be noted that the firing rate mentioned here and later refers to the ratio of the number of spikes to the discrete simulation time.} as the output of SNNs, and convert it to represent continuous action space (i.e., the deterministic policy) through a fully-connected (FC) layer, which bring the floating-point matrix operations to the FC layer, making the whole SNN unable to deploy on the neuromorphic hardware directly.

To develop a fully spiking actor network which could be deployed on neuromorphic chips without any floating-point matrix operations, the key issue is to design a novel neural coding method to decode the spike-train into the continuous value in RL algorithms, such as value estimate or continuous action space, realizing end-to-end spike-based RL. In nature, sensory neurons receive information from the external environment and transmit it to non-spiking interneurons through action potentials~\cite{bidaye2018six}, and then change the membrane voltage of motor neurons through graded signals to achieve effective locomotion. As a translational unit, non-spiking interneurons could affect the motor output according to the sensory input. Inspired by biological research on sensorimotor neuron path, we propose a novel neural coding method to train SNNs for continuous control, where the membrane voltage of non-spiking neurons is used to represent the continuous value in RL algorithms. As shown in Fig. \ref{fig:sensorimotor}, the SNN receives the state from the environment and encodes them as spike-train by spiking neurons. Then, the non-spiking neurons are subsequently introduced to calculate the membrane voltage from the spike-train, which is further used to select the specific action to be performed. Finally, the agent controls the motor to adjust the motion according to the action.

In addition to the membrane voltage coding method, we also integrate the intra-layer connections into the output populations to learn better action representation. The intra-layer connections mainly contain self connections and lateral connections, which widely exist in various brain areas~\cite{satel2013mechanisms}. This mechanism can cause generalization across actions~\cite{fremaux2013reinforcement} and retain the information in the neuron population~\cite{evanusa2019event}, which is applied to the neuron population encoding the same action dimension. Therefore, we propose a fully spiking actor network with intra-layer connections, named ILC-SAN, which is a novel method based on hybrid framework. The experiment results show that our method achieves the start-of-the-art performance on OpenAI gym tasks. The main contributions of this paper are summarized as follows:

\begin{enumerate}
\item{A novel coding method to decode the spike-train for RL algorithms is proposed, which could represent the continuous value by the membrane voltage of non-spiking neurons. The method could be integrated into any RL algorithms, realizing end-to-end spike-based RL. The results of the experiment show that optimal performance could be achieved using the last membrane voltage of non-leaky integrate-and-fire neurons.}
\item{A novel intra-layer connection mechanism is proposed, which has been proved by experiments that it can effectively improve the representation capacity of the output populations.}
\item{A novel spiking actor network (ILC-SAN) is proposed. To our best knowledge, our ILC-SAN is the first fully spiking neural network to achieve the same level of performance as the mainstream DRL algorithms, which ensures that all matrix operations can be completed on the neuromorphic hardware after deploying the trained actor network. The evaluation of continuous control tasks from OpenAI gym demonstrates the effectiveness of ILC-SAN in performance and energy efficiency. Under the same experimental configurations, our ILC-SAN achieves the start-of-the-art performance.}
\end{enumerate}

\section{Related Work}

\subsection{Reward-based Learning by Three-factor Learning Rules}

To bridge the gap between the time scales of behavior and neuronal action potential, modern theories of synaptic plasticity assume that the co-activation of presynaptic and postsynaptic neurons sets a flag at the synapse, called eligibility trace~\cite{sutton2018reinforcement}. Only if a third factor, signaling reward, punishment, surprise, or novelty, exists while the flag is set, the synaptic weight will change. Although the theoretical framework of three-factor learning rules has been developed in the past few decades, experimental evidence supporting eligibility trace has only been collected in the past few years~\cite{gerstner2018eligibility}. Through the derivation of the RL rule for continuous time, the existing approaches have been able to solve the standard control tasks~\cite{fremaux2013reinforcement} and robot control tasks~\cite{mahadevuni2017navigating}. Moreover, Vlasov {\it et al.}~\cite{vlasov2022reinforcement} demonstrate the in-principle possibility to build SNNs with memristor-based synapses trained by inherent local plasticity rules. However, these methods are only suitable for shallow SNNs and low-dimensional control tasks. To solve these problems, Bellec {\it et al.}~\cite{bellec2020solution} propose a learning method for recurrently connected networks of spiking neurons, which is called e-prop. Although the agent learned by reward-based e-prop successfully wins Atari games, the need that the network architecture and numerous hyper-parameters related to neurons and synapses should be manually adjusted between different tasks limits the application of this method.

\subsection{ANN to SNN Conversion for RL}

By matching the firing rate of spiking neurons with the graded activation of analog neurons, trained ANNs can be converted into corresponding SNNs with few accuracy loss~\cite{rueckauer2017conversion}. For the SNNs converted from the ANNs trained by the DQN algorithm, the firing rate of spiking neurons in the output layer is proportional to the Q-value of the corresponding action, which makes it possible to select actions according to the relative size of the Q-value~\cite{patel2019improved,tan2020strategy}. But there is a trade-off between accuracy and efficiency, which tells us that longer inference latency is needed to improve accuracy. As far as we know, for RL tasks, the converted SNNs cannot achieve better results than ANNs.

\subsection{RL Methods using Spike-based BP}

Following the surrogate gradient method~\cite{lee2016training}, the spike-based backpropagation (BP) algorithm has quickly become the mainstream solution for training multi-layer SNNs~\cite{fang2021incorporating,fang2021deep}. Tang {\it et al.}~\cite{tang2020reinforcement} first propose the hybrid framework, composed of a SAN and a deep critic network. Through the co-learning of the two networks, the hybrid framework avoids the problem of value estimation using SNNs. However, the method of scaling the firing rate makes it difficult to accurately characterize the continuous action space, which greatly limits the choice of actions. To improve the representation capacity of SNNs, Tang {\it et al.}~\cite{tang2021deep} propose a population-coded spiking actor network (PopSAN), which achieves the same level of performance as the deep network. Based on this work, a multiscale dynamic coding improved spiking actor network (MDC-SAN)~\cite{zhang2022multiscale} is proposed and performs better than its counterpart deep actor network (DAN). In the decoding stage, the two methods use the fully-connected layer to convert the firing rate into continuous action. Recently, Liu {\it et al.}~\cite{liu2022human} propose a direct spiking learning algorithm for the deep spiking Q-network (DSQN), using a fully-connected layer to decode the firing rate into Q-value (i.e., the state-action value). Sun {\it et al.}~\cite{sun2022solving} implement the Q-value in the same way. Nevertheless, these methods also bring new problems. Since the neuromorphic hardware only accepts the spike input in the form of 0 or 1, matrix operations of decoders need to be completed by other traditional hardware (i.e., CPU, GPU, or embedded AI chip). The resulting energy consumption problem is not reflected in the analysis of previous work~\cite{tang2021deep}, which makes the energy consumption analysis have great defects. Different from these methods, we use the membrane voltage of non-spiking neurons to represent the continuous value in RL algorithms, such as value estimate or continuous action space, so that our method can be directly applied to neuromorphic hardware. Take Loihi as an example, the internal structure of the neuromorphic core mainly includes four primary operating modes: input spike handling, neuron compartment updates, output spike generation, and synaptic updates~\cite{davies2018loihi}. By skipping the process of output spike generation that simulates axons, Loihi supports non-spiking compartments without voltage reset.

Besides, Akl {\it et al.}~\cite{akl2023toward} use the membrane potential readout as the decoding method. However, it lacks a detailed analysis of membrane potential readout methods. Moreover, they encode the observation into a two-neuron input scheme first and then multiply the encoded vector by the weight matrix. Qin {\it et al.}~\cite{qin2022low} use a learnable matrix to compress spike-trains in the temporal dimension through matrix multiplication, which is also a feasible decoding method. Both of these two methods require floating-point matrix multiplication, and cannot be completed only using the neuromorphic hardware. Compared with them, the proposed method employs the population encoder to realize full spike-based RL. In addition, we propose a novel intra-layer connection mechanism to enhance the action decoding.

\subsection{Non-spiking Neurons}

As shown in previous studies~\cite{cramer2020heidelberg,lee2020spike,kheradpisheh2018stdp} and open-source frameworks~\cite{SpikingJelly}, the membrane voltage of non-spiking neurons is feasible to represent a continuous value in spike-based BP methods. However, how to use it to effectively train SNNs for RL has not been systematically studied and remains unsolved, which is a goal of this paper. Moreover, inspired by astrocytes, a Loihi-run central pattern generator (CPG)~\cite{polykretis2020astrocyte} is proposed, which uses non-spiking neurons to generate bursting spikes to control the motor. Their work illustrates the feasibility of non-spiking neurons on Loihi and complements the lack of specific motor control in our work, which constitutes a complete simulation of the sensor motor neuron path (Fig. \ref{fig:sensorimotor}).

\subsection{Intra-layer Connections}

Intra-layer connections play an essential role in the brain and are widely used in SNNs. Generally, intra-layer connections can be divided into self connections and lateral connections. The former is widely used in spiking neural models to provide more complex and adaptive dynamics. The latter, also known as lateral interactions, was previously used to build different receptive fields in visual tasks.

Caused by the reset mechanism, spiking neural models have the characteristics of self connection. Recurrent Spiking Neural Network (RSNN) fully exploits the self connection characteristics of spiking neural models, making them more adaptive~\cite{bellec2018long,bellec2020solution,yin2021accurate}. However, most of them are based on soft reset or hard reset, and cannot describe hyperpolarization, which reduces the readiness of biological neurons to fire again after recent firing activity. To solve this problem, AHP-neurons~\cite{rao2022long} model the after-hyperpolarizing (AHP) currents, leading to spike frequency adaptation. Although the update to the AHP-current in response to an output spike is fixed by a hyper-parameter, AHP-neurons develop a new way of self connection and provide a principled alternative to LSTM units for sequence processing tasks.

In the field of neuroscience, the work on lateral interactions mainly focuses on the retina~\cite{ratliff1974dynamics}, which makes most of the work on lateral interactions in SNNs related to visual tasks~\cite{diehl2015unsupervised,evanusa2019event,cheng2020lisnn}. Because of its ability to sharpen edges and highlight contours, lateral interactions are used to extract features from images. Furthermore, lateral interactions are widespread in various regions of the brain~\cite{satel2013mechanisms}. Fr{\'e}maux {\it et al.}~\cite{fremaux2013reinforcement} use population coding to represent actions, and each neuron encodes a different motion vector. They adopt fixed lateral interactions in the neuron population to implement the N-winner-take-all mechanism, leading to generalization across actions. However, there are several problems in this method, such as depending on the relatively low-dimension action space, a large number of hyper-parameters need to be adjusted for different action spaces, which limits its application in complex tasks.

Zhang {\it et al.}~\cite{zhang2019spike} use a method similar to ours to model intra-layer connections. The difference is that the spike-train level post-synaptic potentials (S-PSPs) are transmitted between neurons from the same layer, rather than the spike signals in our method. However, S-PSPs rely on a relatively long spike-train, which makes its simulation time much longer than the methods using back-propagation through time (e.g. ours)~\cite{fang2021incorporating}, resulting in relatively high energy consumption and inference time.

\begin{figure*}[t]
\centering
\includegraphics[width=2\columnwidth]{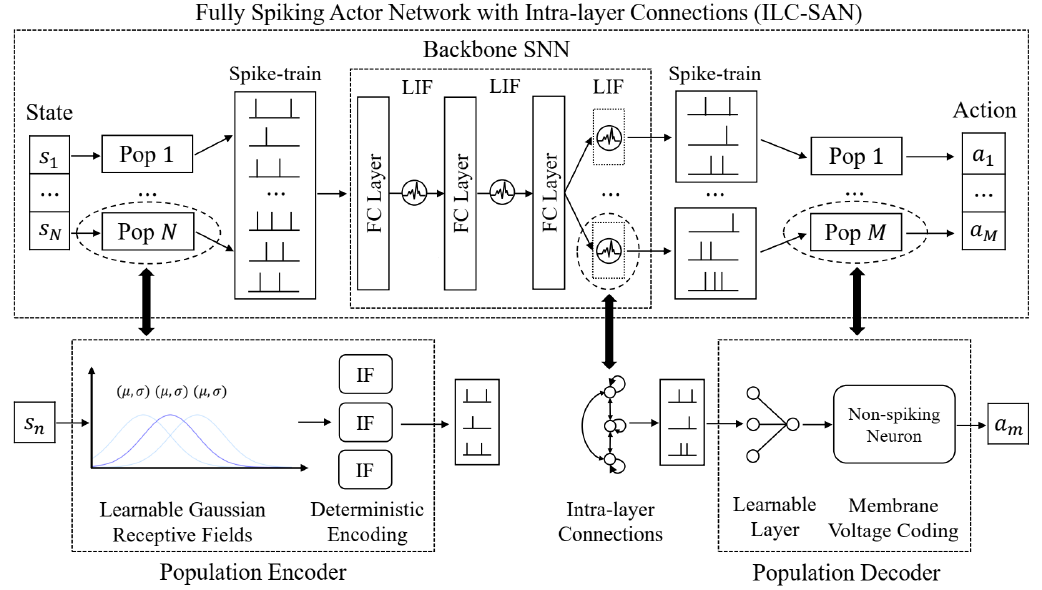}
\caption{The overall framework of the proposed ILC-SAN. The state is transformed into spike-trains by the population encoder. Each state dimension $s_n$ is encoded by the corresponding input population, which consists of learnable Gaussian receptive fields and Integrate-and-Fire (IF) neurons. Each neuron in the input population has a different Gaussian kernel ($\mu, \sigma$). Through these Gaussian kernels, $s_n$ is first encoded into the stimulation strength for each neuron in the input population, and then transformed into the spike-trains using deterministic encoding. After that, the spike-trains are transmitted through the backbone SNN to the population decoder. The spiking neurons in the last layer of the backbone SNN are evenly divided into $M$ output populations, and the intra-layer connections are applied in each output population. Each output population has a corresponding population decoder, where the spike-trains are first integrated into a single non-spiking neuron and then decoded into the corresponding action dimension using membrane voltage coding.}
\label{fig:san}
\end{figure*}

\section{Method}
In this paper, we focus on the typical control task where the agent needs to learn multi-dimensional deterministic policies. Note the RL methods~\cite{lillicrap2015continuous,fujimoto2018addressing,haarnoja2018soft} in this field often utilize the actor-critic framework, where the critic network is just used for training, and only the actor network is deployed on the hardware of the application. Hence, our method aims to develop a fully spiking actor network that could be deployed on neuromorphic chips without any floating-point matrix operations. In this section, we first introduce the spiking neural model and its discrete dynamics. Then, we propose the non-spiking neurons and analyze the membrane voltage coding. Finally, we present the implemented details of our ILC-SAN (Fig. \ref{fig:san}), which can be trained in conjunction with a deep critic network using the DRL algorithms. Moreover, the trained ILC-SAN provides an energy-efficient solution for continuous control tasks due to its fully spiking architecture.

\begin{figure}[t]
\centering
\includegraphics[width=\columnwidth]{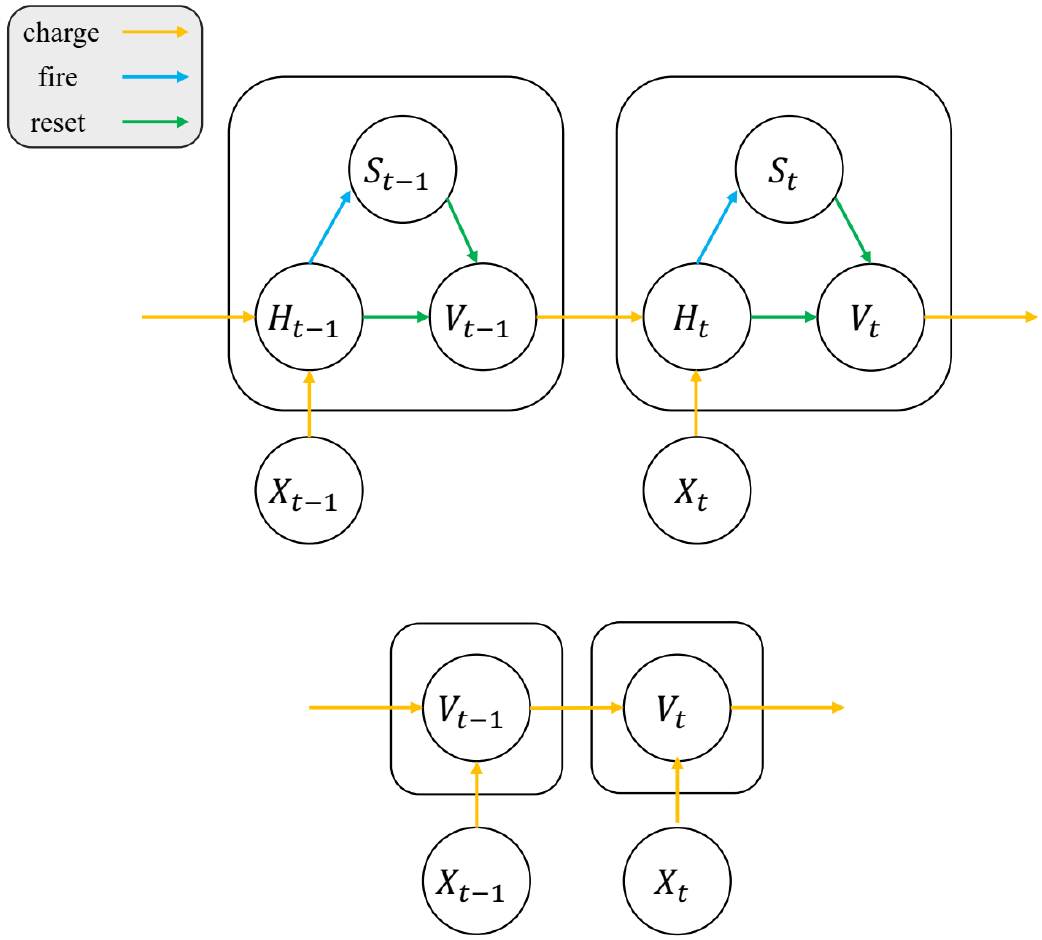} 
\caption{The general discrete neural model (Top) Spiking neural model. (Bottom) Non-spiking neural model.}
\label{fig:neural_model}
\end{figure}

\subsection{Spiking Neural Model}

The basic computing unit of SNNs is the spiking neuron. In the open-source deep learning framework for SNNs, spiking neural models are usually simulated in discrete time-steps. The dynamics of most kinds of discrete spiking neurons can be described as:
\begin{equation}
\label{eq:charging}
    H_t=f\left(V_{t-1}, X_t\right),
\end{equation}
\begin{equation}
\label{eq:firing}
    S_t=\mathrm{\Theta}\left(H_t-V_{th}\right),
\end{equation}
\begin{equation}
\label{eq:resetting}
    V_t=H_t(1-S_t)+V_{reset}S_t,
\end{equation}
where $H_t$ and $V_t$ denote the membrane voltage after neural dynamics and the trigger of a spike at time-step $t$, respectively. $X_t$ denotes the external input, and $S_t$ means the output spike at time-step $t$, which equals 1 if there is a spike and 0 otherwise. $V_{th}$ denotes the threshold voltage and $V_{reset}$ denotes the membrane reset voltage. As shown in Fig. \ref{fig:neural_model}, Eq. (\ref{eq:charging}) - (\ref{eq:resetting}) establish a general mathematical model to describe the discrete dynamics of spiking neurons, which include charging, firing and resetting. The dynamics described by the Eq. (\ref{eq:resetting}) are called the hard reset. In addition, spiking neurons can also adopt soft reset, whose dynamics can be described as:
\begin{equation}
\label{eq:soft_reset}
    V_t=H_t-V_{th}S_t.
\end{equation}
Specifically, Eq. (\ref{eq:charging}) describes the subthreshold dynamics, which vary with the type of neuron models. Here we consider the Integrate-and-Fire (IF) model~\cite{brunel2007lapicque} and the Leaky Integrate-and-Fire (LIF) model~\cite{gerstner2014neuronal}, the two most commonly used spiking neuron models. The function $f(\cdot)$ of the LIF neuron is defined as:
\begin{equation}
\label{eq:LIF}
    f\left(V_{t-1},X_t\right)=\alpha_V(V_{t-1}-V_{reset})+V_{reset}+X_t,
\end{equation}
where $\alpha_V$ is the voltage decay factor. When $\alpha_V=1$, this equation represents the function $f(\cdot)$ of the IF neuron. The spike generative function $\Theta(x)$ is the Heaviside step function, which is defined by $\Theta(x)=1$ for $x\geq 0$ and $\Theta(x)=0$ for $x<0$. Note that $V_0=V_{reset}$, $S_0=0$.

\subsection{Non-spiking Neural Model}

Non-spiking neurons can be regarded as a special case of spiking neurons. If we set the threshold voltage $V_{th}$ of spiking neurons to infinity, the dynamics of neurons will always be under the threshold, which is so-called non-spiking neurons. Therefore, the application of non-spiking neurons does not affect our network to be a fully spiking neural network. Since non-spiking neurons do not have the dynamics of firing and resetting, we could simplify the neural model to the following equation (see Fig. \ref{fig:neural_model}):
\begin{equation}
\label{eq:nonspiking}
    V_t=f\left(V_{t-1},X_t\right).
\end{equation}

Here we consider the non-spiking LIF model, which can also be called the Leaky Integrate (LI) model. The dynamics of LI neurons are described by the following equation:
\begin{equation}
\label{eq:integrate}
    V_t=\alpha_V(V_{t-1}-V_{reset})+V_{reset}+X_t.
\end{equation}
When $\alpha_V=1$, this equation represents the Integrate model.

\subsection{Membrane Voltage Coding}

According to the definition of SNNs, the output is a spike-train. However, the output of the functions in RL algorithms are all continuous values. To bridge the difference between these two data forms, we need a spike decoder to complete the data conversion. Inspired by the non-spiking interneurons found in insects, we propose to use non-spiking neurons as a bridge between perception and motion for decision-making. In the sensorimotor neuron path, spike signals transmitted by sensory neurons are integrated into non-spiking neurons. Then the agents use the membrane voltage of non-spiking neurons to make decisions, which determines the input current of motor neurons.

In the whole simulation time $T$, the non-spiking neurons take the spike-train as the input sequence, and then the membrane voltage $V_t$ at each time-step $t$ can be obtained. To represent the final output $O$, we need to choose an optimal statistic according to the membrane voltage at all times, i.e., $O=Stat(V_1, \cdots, V_T)$.

To meet the needs of the algorithm, we finally design three statistics as candidates: 
\begin{itemize}
    \item \textbf{Last membrane voltage} $D_{last}$: It is a natural idea to use the last membrane voltage after the simulation time as the characterization to make full use of all simulations. The formula of $D_{last}$ is as follows:
    \begin{equation}
        O=V_T.
    \end{equation}
    \item \textbf{Membrane voltage with maximum absolute value} $D_{max}$: By recording the membrane voltage of non-spiking neurons at each time-step in the whole simulation time, we can get the sequence of the membrane voltage. In previous work~\cite{cramer2020heidelberg}, the maximum membrane voltage is used to represent the probability of each category. However, this statistic is more suitable for representing non-negative numbers. Due to the unknown sign of continuous values, the membrane voltage with the maximum absolute value is undoubtedly a better statistic. The formula of $D_{max}$ is as follows:
    \begin{equation}
        O=V_{\mathop{\arg\max}\limits_{1\le t\le T}{\lvert V_t \rvert}}.
    \end{equation}
    \item \textbf{Mean membrane voltage} $D_{mean}$: Similar to the maximum membrane voltage, we can obtain the mean value by collecting the membrane voltage at each time-step, which is also a meaningful statistic. The formula of $D_{mean}$ is as follows:
    \begin{equation}
        O=\frac{1}{T}\sum_{t=1}^{T}V_t.
    \end{equation}
\end{itemize}

For these statistics, we empirically evaluate them in experiments respectively in Section \ref{sec:analysis}. We find both $D_{last}$ and $D_{max}$ are effective and $D_{last}$ performs better. In addition, it is obvious that $D_{last}$ is easier to implement and deploy than $D_{max}$. Hence, we choose $D_{last}$ as the default membrane voltage coding method in the following sections. It should be noted that the final output $O$ in this paper represents the action of any dimension.

\subsection{Fully Spiking Actor Network with Intra-layer Connections}

In this subsection, we present the implemented details of our ILC-SAN. With the help of the membrane voltage of non-spiking neurons, we incorporate all matrix operations of SAN into the neuromorphic chips, solving the problem of floating-point matrix operations in the decoding stage. By applying intra-layer connections in the output populations, the learned action representation could be better.

\subsubsection{Population Encoder}

For the $N$-dimensional state $\bm{s}\in \mathbb{R}^N$, we use a population of neurons $E$ with different Gaussian receptive fields ($\bm{\mu}, \bm{\sigma}$) to encode each state dimension $s_i$ into a spike-train, where $\bm{\mu}$ and $\bm{\sigma}$ are task-specific trainable parameters. Suppose $P_{in}$ represents the input population sizes per state dimension, the shape of the spike-train $\bm{ST}$ is ($N\cdot P_{in}, T$). The calculation process of the encoder can be divided into two stages. First, the state $\bm{s}$ is converted to the stimulation strength of each neuron in the population $\bm{A}_E$:
\begin{equation}
\label{eq:pop}
    \bm{A}_E=e^{-\frac{1}{2}(\frac{s_i-\bm{\mu}}{\bm{\sigma}})^2}.
\end{equation}
Second, the computed $\bm{A}_E$ is used to generate the spike-train. For performance reasons~\cite{tang2021deep}, we use deterministic encoding, where the neurons are simulated as soft-reset IF neurons and $\bm{A}_E$ is the external input ($\bm{X}_t=\bm{A}_E$). Therefore, we call this population encoder $E_{pop\_det}$ for short. For simplicity, we directly back-propagate the gradient w.r.t. the stimulation strength $\bm{A}_E$ regardless of whether a spike is fired or not at any time-step $t$, $\frac{\delta \bm{S}_t}{\delta \bm{A}_E}=1$.

\subsubsection{Current-based LIF Neurons}

For a fair comparison, we employ the current-based LIF (CLIF) model of spiking neurons, which is  used in previous work~\cite{tang2021deep}. The CLIF neurons can integrate presynaptic spikes into the current and subsequently integrate the current into the membrane voltage ($\bm{X}_t=\bm{C}_t$), the detailed dynamics of which are described in Algorithm \ref{alg:alg1}. $\alpha_C$ and $\alpha_V$ are the current and voltage decay factors. To distinguish from the voltage decay factors of non-spiking neurons, we express them as $\alpha_V^{LI}$ and $\alpha_V^{CLIF}$. And the membrane reset voltage also uses different superscripts to distinguish the symbols. Except for the population encoder, all spiking neural models of ILC-SAN adopt hard reset. The rectangular function is used as the surrogate function. The gradient is defined by:
\begin{equation}
    \Theta^\prime(x)=\left\{\begin{matrix}1&\lvert x \lvert<w\\0&otherwise\\\end{matrix}\right.,
\end{equation}
where $w$ is the threshold window for passing the gradient.

\subsubsection{Backbone SNN}

The backbone SNN connects the population encoder and the population decoder, whose input and output are spike-trains. The typical architecture of SNNs consists of synaptic layers and neuronal layers~\cite{fang2021incorporating}. The synaptic layers include convolutional layers and fully-connected layers, each of which is followed by a neuronal layer. Since the population encoder will bring extremely high computation costs when handling high-dimensional data such as images or spatiotemporal signals, we extract the state feature as a 1-D vector and only use the fully-connected layer as the synaptic layer. Moreover, the CLIF neurons form the neuronal layers. In addition, all weight parameters are shared at all simulation time-steps. Note that we treat the synaptic layer and its subsequent neuronal layer as one layer in formula derivation. The former plays a similar role to dendrites in neuronal cells, and the latter works like the bodies and axons. For a task with $M$-dimensional actions, we equally divide the spiking neurons of the last layer into $M$ output population with a size of $P_{out}$. Each output population has a corresponding population decoder. Suppose that the backbone SNN has $L$ layers, $\bm{W}^l$, $\bm{b}^l$ represent the weights and biases of the $l$-th layer. $\bm{C}_t^l$ represents the current for CLIF neurons of the $l$-th layer at time-step $t$. Similarly, $\bm{H}_t^l$, $\bm{V}_t^l$, $\bm{S}_t^l$ represent different variables in CLIF neurons of the $l$-th layer at time-step $t$. Note that we use $\bm{S}_t^0$ to represent the spike input of the backbone SNN at time-step $t$, i.e., the output of population encoder at time-step $t$, $\bm{ST}_t$.

\subsubsection{Population Decoder}

Each population decoder consists of a learnable layer and an LI neuron. The $m$-th population decoder projects the spike-trains from the $m$-th output population into the $m$-th action dimension, $m\in\{1, \cdots, M\}$. Specifically, in the $m$-th population decoder, the spike-trains from the $m$-th output population are inputted to the learnable layer first, and then the membrane voltage of the LI neuron is updated at each time-step $t$. After every $T$ simulation time-steps, the input of the $m$-th population decoder is decoded into the specified statistic of membrane voltage $O^m$, which represents the value of the $m$-th action dimension. For the $m$-th output population, $\bm{W}_D^m$, $b_D^m$ represent the weights and biases of the decoder, and $X_t^m$, $V_t^m$ represent different variables in LI neurons at time-step $t$.

\subsubsection{Intra-layer Connections}

For continuous control tasks from OpenAI gym, actions in different dimensions vary greatly. Therefore, we apply intra-layer connections for each output population, that is, intra-layer connections only occur between neuron populations encoding the same dimension of continuous action space. For the $m$-th output population, the current received from the intra-layer connections at the next time-step can be calculated as follows:
\begin{equation}
\label{eq:intra}
    \bm{I}_{t+1}^{m}=\bm{W}_{intra}^m\bm{S}_t^{L,m}+\bm{b}_{intra}^m,
\end{equation}
where $\bm{W}_{intra}^m$, $\bm{b}_{intra}^m$ represent the weights and biases of the intra-layer connections. $\bm{S}_t^{L,m}$ represent the spike output of the backbone SNN used in the $m$-th output population at the time-step $t$. Note that $\bm{W}_{intra}^m$ is a square matrix of order $P_{out}$. Since the intra-layer connection and the inter-layer connection both are connections between neurons, we model them in the same way as fully-connected layers. As similar to the implementation of RNNs\footnote{https://pytorch.org/docs/stable/generated/torch.nn.RNN.html.}, $\bm{b}_{intra}^m$ is introduced to improve the fitting ability of $\bm{W}^m_{intra}\bm{S}_t^{L,m}$ by learning the data bias.

\begin{algorithm}[t]
\caption{Forward propagation through ILC-SAN}\label{alg:alg1}
\textbf{Input}: $N$-dimensional state $\bm{s}$\\
\textbf{Output}: $M$-dimensional action $\bm{a}$

\begin{algorithmic}[]
\STATE Initialize encoding means $\bm{\mu}$ and standard deviations $\bm{\sigma}$ for the population encoder;
\STATE Randomly initialize the $L$-layer backbone SNN;
\STATE Randomly initialize the population decoder;
\STATE Compute the spike-train from input populations generated by the encoder: $\bm{ST}=Encoder(\bm{s}, \bm{\mu}, \bm{\sigma})$;
\FOR{$t=1, \cdots, T$}
    \STATE Spikes from input populations at time-step $t$: $\bm{S}_t^0=\bm{ST}_t$;
    \FOR{$l=1, \cdots, L$}
        \STATE Update CLIF neurons in layer $l$ at time-step $t$:
        \IF{$l=L \land t\neq 1$}
            \STATE Integrate spikes from layer $l-1$ and layer $l$ into the current:
            \STATE $\bm{C}_t^l=\alpha_C\cdot \bm{C}_{t-1}^l+\bm{W}^l\bm{S}_t^{l-1}+\bm{b}^l+\bm{I}_{t}$;
        \ELSE
            \STATE Integrate spikes from layer $l-1$ into the current:
                \STATE $\bm{C}_t^l=\alpha_C\cdot \bm{C}_{t-1}^l+\bm{W}^l\bm{S}_t^{l-1}+\bm{b}^l$;
        \ENDIF
        \STATE $\bm{H}_t^l=\alpha_V^{CLIF}(\bm{V}_{t-1}^l-V_{reset}^{CLIF})+V_{reset}^{CLIF}+\bm{C}_t^l$;
        \STATE $\bm{S}_t^l=\mathrm{\Theta}(\bm{H}_t^l-V_{th})$;
        \STATE $\bm{V}_t^l=\bm{H}_t^l(1-\bm{S}_t^l)+V_{reset}^{CLIF}\bm{S}_t^l$;
    \ENDFOR
    \STATE Divide $\bm{S}_t^{L}$ evenly among $M$ output populations: $\{\bm{S}_t^{L,m}\},\ m=1, \cdots, M$;
    \FOR{$m=1, \cdots, M$}
        \IF{$t\neq T$}
            \STATE Calculate the current generated by intra-layer connections $\bm{I}_{t+1}^{m}$ using Eq. (\ref{eq:intra});
        \ENDIF
        \STATE Update the LI neuron of the $m$-th output population:
        \STATE $X_t^m=\bm{W}_D^m\bm{S}_t^{L,m}+b_D^m$;
        \STATE $V_t^m=\alpha_V^{LI}(V_{t-1}^m-V_{reset}^{LI})+V_{reset}^{LI}+X_t^m$;
    \ENDFOR
    \IF{$t\neq T$}
        \STATE Merge $\{\bm{I}_{t+1}^{m}\}$, $m=1, \cdots, M$ into $\bm{I}_{t+1}$;
    \ENDIF
\ENDFOR
\STATE Generate $M$-dimensional action $\bm{a}$: 
\STATE $a_m=O^m=Stat(V_1^m, \cdots, V_T^m)$, $m=1, \cdots, M$;
\end{algorithmic}
\end{algorithm}

\subsection{ILC-SAN embedded into TD3}

Our ILC-SAN is functionally equivalent to a DAN, which can be trained in conjunction with a deep critic network using TD3 algorithms~\cite{fujimoto2018addressing}. During training, the ILC-SAN establishes a mapping between states and actions to represent the policy of agents. The deep critic network estimates the corresponding Q-value, which can guide the ILC-SAN to learn a better policy. During the evaluation, the trained ILC-SAN can predict the action with the maximum Q-value produced by the trained critic network. More details can be found in the forward propagation of ILC-SAN (Algorithm \ref{alg:alg1}).

\section{Experiments}
\label{sec:exp}

In this section, we first evaluate the performance of ILC-SAN on eight continuous control tasks from OpenAI gym. Then we analyze the effects of different decoders on performance, and evaluate the influence of each component in the intra-layer connections. Finally, we compare the theoretical energy consumption of different models according to the power efficiency of their advanced hardware. In addition, our experiments are built upon the open-source code of PopSAN~\cite{tang2021deep}.

\begin{figure}[t]
\centering
\subfloat[]{\includegraphics[width=0.24\linewidth]{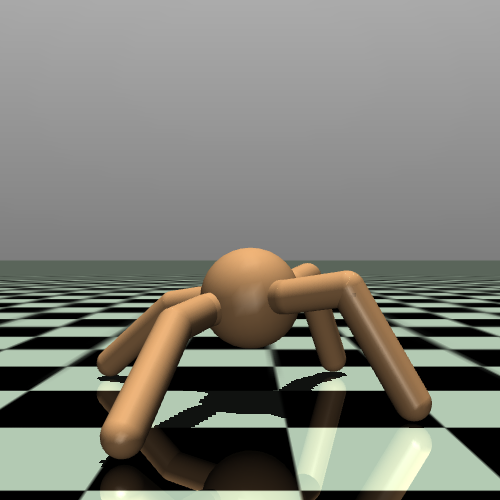}}
\subfloat[]{\includegraphics[width=0.24\linewidth]{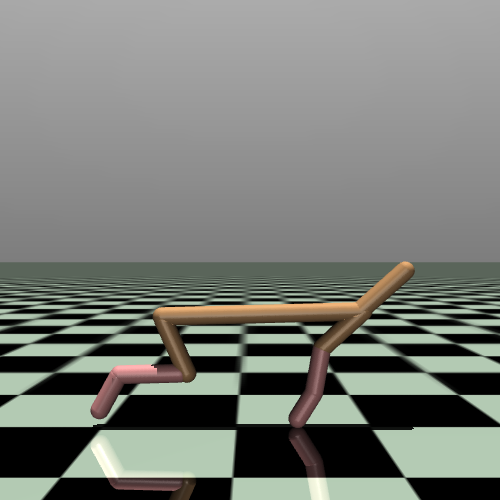}}
\subfloat[]{\includegraphics[width=0.24\linewidth]{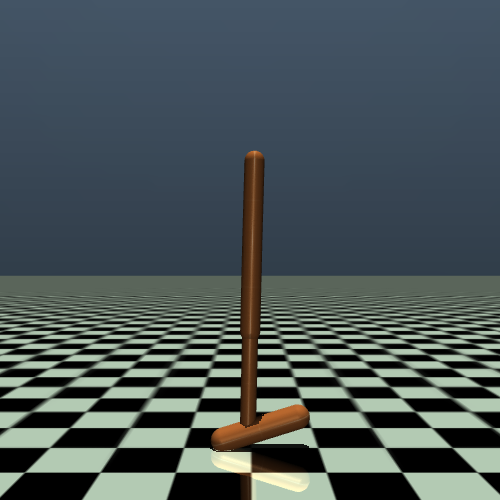}}
\subfloat[]{\includegraphics[width=0.24\linewidth]{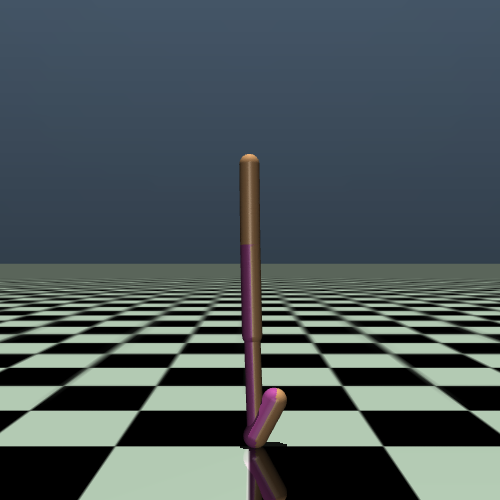}}

\subfloat[]{\includegraphics[width=0.24\linewidth]{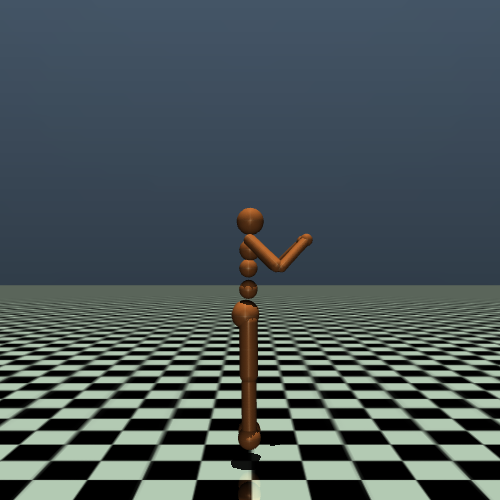}}
\subfloat[]{\includegraphics[width=0.24\linewidth]{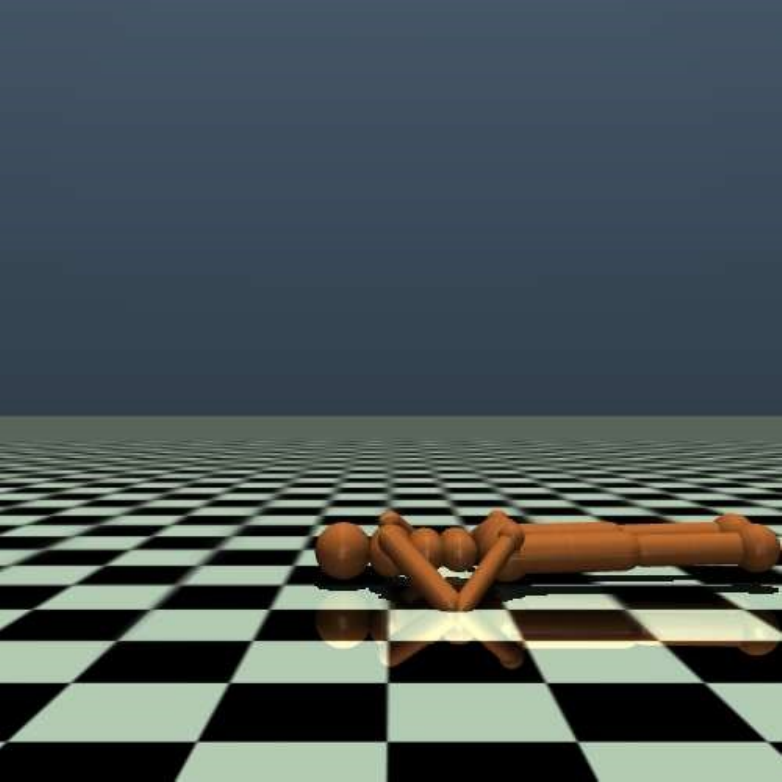}}
\subfloat[]{\includegraphics[width=0.24\linewidth]{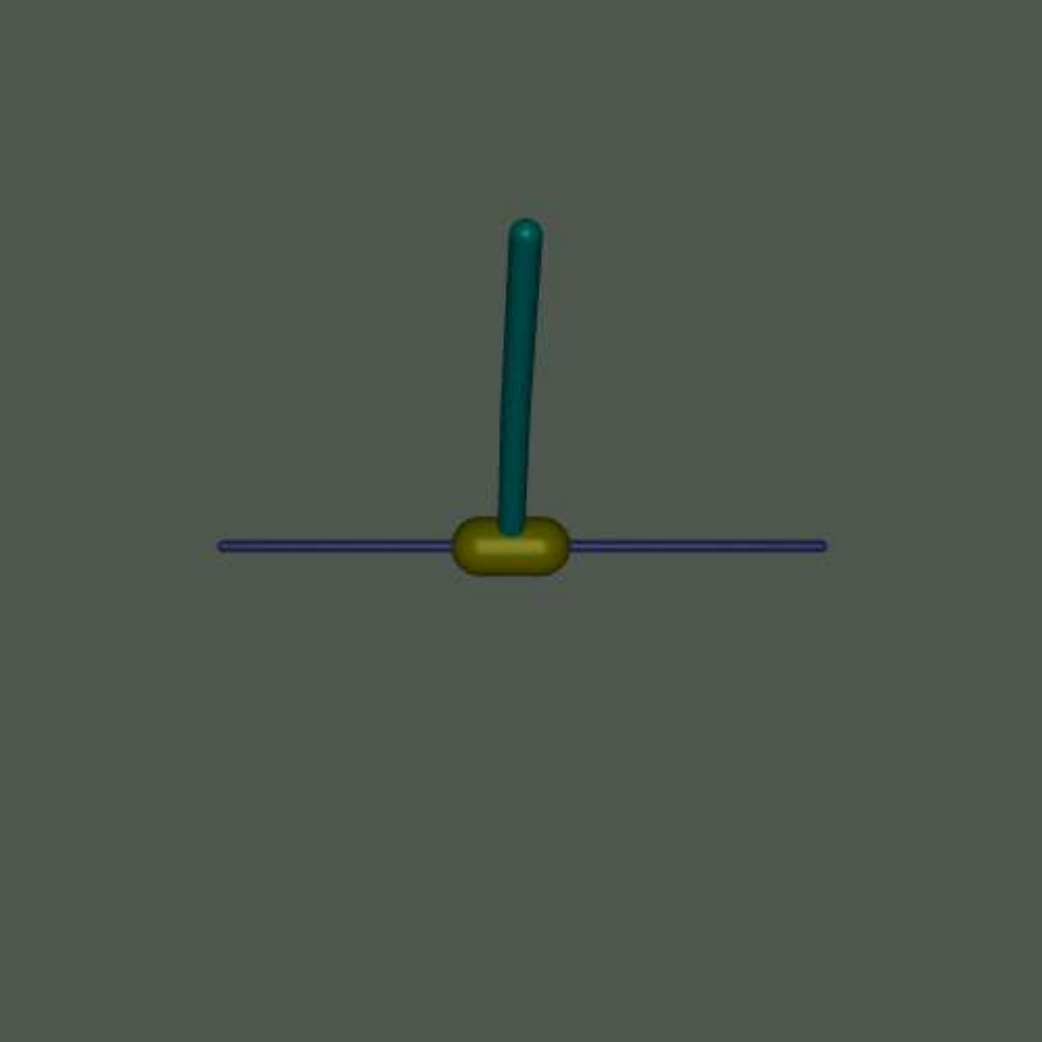}}
\subfloat[]{\includegraphics[width=0.24\linewidth]{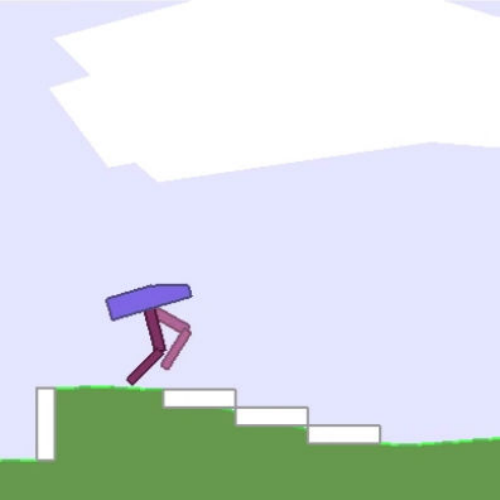}}
\caption{Eight continuous control tasks from OpenAI gym.} (a) Ant-v3: make a 3D four-legged robot move forward as fast as possible by applying torques on the eight hinges (b) HalfCheetah-v3: make a 2D robot run forward as fast as possible by applying a torque on the joints (c) Hopper-v3: make a 2D one-legged figure hop forward as fast as possible by applying torques on the three hinges (d) Walker2d-v3: make a 2D two-legged figure move forward as fast as possible by applying torques on the six hinges. (e) Humanoid-v3: make a 3D bipedal robot walk forward as fast as possible without falling over by applying torques on the seventeen hinges. (f) HumanoidStandup-v2: make a 3D bipedal robot stand up and then keep it standing by applying torques on the seventeen hinges. (g) InvertedDoublePendulum-v2: balance the second pole on top of the first pole by applying continuous forces on the cart. (h) BipedalWalker-v3: make a 2D walker robot walk forward without falling over by controlling the motor speed for each of the four joints at both hips and knees.
\label{fig:task}
\end{figure}

\subsection{Experimental Settings}

We evaluate our approach on the OpenAI gym tasks, which are often used as a benchmark for continuous control algorithms. All the tasks used are shown in the Fig. \ref{fig:task}. To save computational resources, we only use all the tasks for the comparative experiments in Section \ref{sec:comparison}, while for other experiments, we use the first four most commonly used tasks. To ensure the reproducibility, each of our models is trained for ten rounds, corresponding to ten random seeds. In each round, the task is trained for 1 million steps and evaluated every 10k steps, where each evaluation reported the average reward over 10 episodes using the deterministic policy. Each episode can last up to 1000 execution steps. In all experiments, we use TD3 algorithm to train the ILC-SAN.

Hyper-parameter configurations for the methods used subsequently are as follows: the deep critic network is ($N+M$, 256, relu, 256, relu, 1); the DAN is ($N$, 256, relu, 256, relu, $M$, tanh); the ILC-SAN is ($N\cdot P_{in}$, 256, CLIF, 256, CLIF, $M\cdot P_{out}$), the population size for each state and action dimension is 10; the learning rate of the DAN and the deep critic network is 1e-3; the learning rate of the ILC-SAN is 1e-4; the reward discount factor is 0.99; the Gaussian exploration noise with stddev is 0.1; the Gaussian smoothing noise for target policy with stddev is 0.2; the maximum length of replay buffer is 1e6; the soft target update factor is 0.005; the batch size is 100; the noise clip is 0.5; the policy delay factor is 2.

For CLIF neurons, we use the same hyper-parameters as the open-source code of PopSAN, so the membrane reset voltage $V_{reset}^{CLIF}$ is 0.0, the threshold voltage $V_{th}$ is 0.5, the current decay factor $\alpha_{C}$ is 0.5, the voltage decay factor $\alpha_{V}^{CLIF}$ is 0.75, and the threshold window $w$ is 0.5. For LI neurons, we set $V_{reset}^{LI}$ to 0.0. Since $\alpha_{V}^{CLIF}$ is a fixed value in all experiments, we use $\alpha_V$ to represent the voltage decay factor of non-spiking neurons in the rest of Section \ref{sec:exp}.

\begin{table}[t]
\caption{State dimension and action dimension of different tasks}
\label{tab:task}
\centering
\begin{tabular}{ccc}
\toprule
Task                      & N   & M \\
\midrule
Ant-v3                    & 111  & 8 \\
HalfCheetah-v3            & 17   & 6 \\
Hopper-v3                 & 11   & 3 \\
Walker2d-v3               & 17   & 6 \\
Humanoid-v3               & 376  & 17 \\
HumanoidStandup-v2        & 376  & 17 \\
InvertedDoublePendulum-v2 & 11   & 1 \\
BipedalWalker-v3          & 24   & 4 \\
\bottomrule
\end{tabular}
\end{table}

\begin{table*}[t]
\caption{Max average rewards over 10 random seeds for DAN, ILC-SAN and other SANs}
\label{tab:perf}
\centering
\resizebox{2\columnwidth}{!}{
\begin{tabular}{ccccccccc}
\toprule
\multirow{2}{*}{Task} & \multirow{2}{*}{DAN~\cite{fujimoto2018addressing}} & \multirow{2}{*}{PopSAN~\cite{tang2021deep}} & \multirow{2}{*}{MDC-SAN~\cite{zhang2022multiscale}} & \multirow{2}{*}{Akl et al.~\cite{akl2023toward}} & \multirow{2}{*}{AC-BCQ~\cite{qin2022low}} & \multicolumn{3}{c}{ILC-SAN} \\
\cline{7-9}
 & & & & & & $E_{layer}$ & $E_{pop}$ & $E_{pop\_det}$ \\
\midrule
Ant-v3               & 5472$\pm$653 & 4848$\pm$1023 & 5410$\pm$748 & 3820$\pm$942 & 4070$\pm$34 & 5752$\pm$258 & 5387$\pm$529 & 5653$\pm$533 \\
HalfCheetah-v3       & 10471$\pm$1695 & 10523$\pm$1142 & 10472$\pm$1146 & 3220$\pm$1857 & 10610$\pm$63 & 10120$\pm$1864 & 10422$\pm$909 & 10376$\pm$1163 \\
Hopper-v3            & 3520$\pm$105 & 517$\pm$1057 & 3554$\pm$153 & 2713$\pm$1424 & 2527$\pm$455 & 3518$\pm$182 & 3547$\pm$163 & 3583$\pm$86 \\
Walker2d-v3          & 4999$\pm$842 & 4199$\pm$1426 & 4604$\pm$809 & 3208$\pm$1514 & 2551$\pm$932 & 5263$\pm$869 & 4919$\pm$627 & 4511$\pm$462 \\
Humanoid-v3          & 3681$\pm$2258 & 5821$\pm$440 & 5879$\pm$225 & 135$\pm$123 & $-$ & 5872$\pm$236 & 5241$\pm$1633 & 5874$\pm$395 \\
HumanoidStandup-v2   & 129271$\pm$13021 & 155387$\pm$21632 & 149815$\pm$10404 & 73620$\pm$33564 & $-$ & 156062$\pm$18252 & 161889$\pm$31302 & 160974$\pm$22132 \\
InvertedDoublePendulum-v2 & 8425$\pm$2950 & 9327$\pm$3 & 9323$\pm$6 & 9338$\pm$8 & $-$ & 9335$\pm$10 & 9336$\pm$13 & 9327$\pm$5 \\
BipedalWalker-v3     & 194$\pm$146 & 267$\pm$93 & 267$\pm$98 & 190$\pm$86 & $-$ & 202$\pm$151 & 272$\pm$100 & 301$\pm$10 \\
\midrule
APR & 100.00\% & 101.81\% & 114.48\% & 63.90\% & 74.63\% & 112.77\% & 114.47\% & \textbf{118.05\%} \\
\bottomrule
\end{tabular}}
\end{table*}

\begin{figure*}[t]
\centering
\includegraphics[width=2\columnwidth]{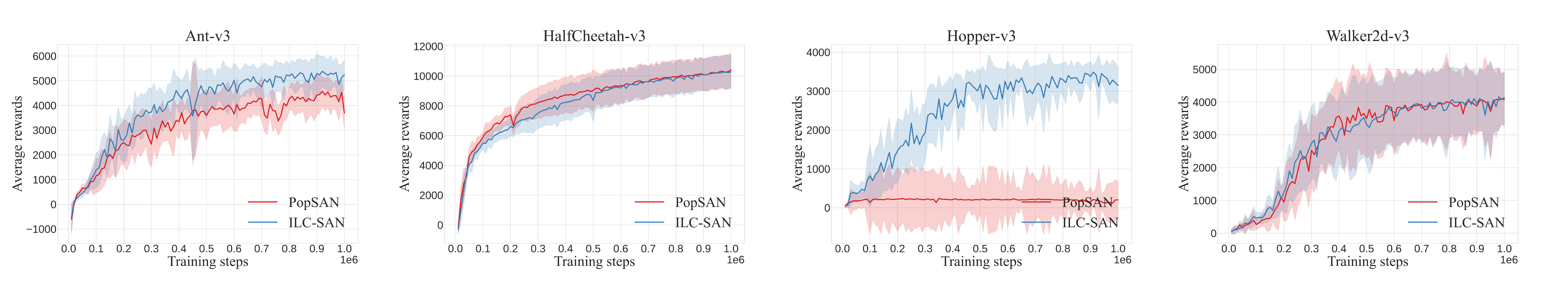}
\caption{The comparison of average rewards for PopSAN and ILC-SAN using $E_{pop\_det}$ over 10 random seeds. The shaded area represents half the value of the standard deviation, and the curves are smoothed for clarity.}
\label{fig:pop-det}
\end{figure*}

\begin{figure*}[t]
\centering
\includegraphics[width=2\columnwidth]{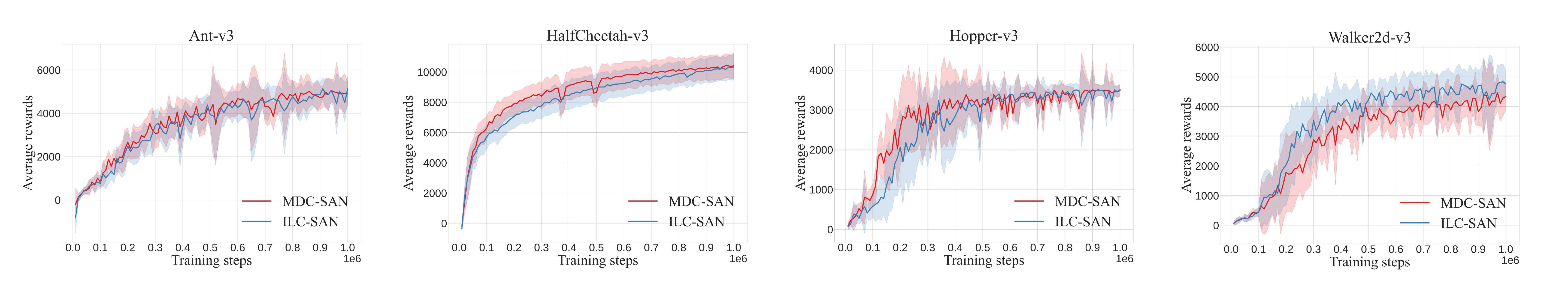}
\caption{The comparison of average rewards for MDC-SAN and ILC-SAN using $E_{pop}$ over 10 random seeds. The shaded area represents half the value of the standard deviation, and the curves are smoothed for clarity.}
\label{fig:pop}
\end{figure*}

\subsection{Comparison with the State-of-the-Art}
\label{sec:comparison}

According to the input of SNNs, we can divide the existing methods into two categories, namely, spike input~\cite{tang2021deep} and floating-point input~\cite{zhang2022multiscale,akl2023toward,qin2022low}. The former considers both energy consumption and performance to facilitate the deployment of neuromorphic hardware. And the latter focuses on performance improvement, so its performance is usually considered higher~\cite{zhang2022multiscale}. When comparing with such methods, we need to modify our population encoder $E_{pop\_det}$. Here we tested two alternative input coding methods, raw population coding $E_{pop}$~\cite{zhang2022multiscale} and learnable layer coding $E_{layer}$~\cite{akl2023toward,qin2022low}. For $E_{pop}$, We remove the second stage of the population encoder and use $\bm{A}_E$ calculated by Eq. (\ref{eq:pop}) as the input of the backbone SNN directly at each time-step $t$, i.e. $\bm{S}_t^0=\bm{A}_E$. For $E_{layer}$, we remove the population encoder and take the state $\bm{s}$ as the input directly at each time-step $t$, i.e. $\bm{S}_t^0=\bm{s}$. Therefore, the size of $\bm{S}_t^0$ is $N\cdot P_{in}$ for $E_{pop\_det}$ and $E_{pop}$, $N$ for $E_{layer}$. Note that only the full method $E_{pop\_det}$ meets the requirements of fully spiking neural networks. $E_{pop}$ and $E_{layer}$ are only used to compare with other methods with the same input coding scheme.

We compare the performance of our ILC-SANs with other SANs, taking the average performance ratio (APR) of different SANs to the corresponding DANs across all the tasks as the measurement standard. This measure can be described as the following equation:
\begin{equation}
    APR=\frac{1}{N_{\mathcal T}}\sum_{task\in{\mathcal T}}\frac{AN_{task}}{DAN_{task}},
\end{equation}
where $\mathcal T$ represents a set of tasks, and $N_{\mathcal T}=|\mathcal T|$. $AN_{task}$ is the max average rewards of the given actor network over 10 random seeds on the corresponding task, where $AN$ can be any actor network, such as DAN, PopSAN, MDC-SAN and ILC-SAN.

To avoid the impact of software package versions and random seeds, almost all the experiments are under the same experimental setting, except for AC-BCQ~\cite{qin2022low}. Since AC-BCQ has not been officially published and its source code has not been released, we directly use the data provided in the paper. We re-run DAN, PopSAN and the method proposed by Akl et al.~\cite{akl2023toward} using their open-source code. Using the best dynamic parameters of dynamic neurons given by the authors, we reproduce the MDC-SAN and re-run it. As Table \ref{tab:perf} shows, our models achieve better performance than PopSAN on all three different input coding methods, which demonstrates that our method has a marked effect for different encoders. When considering ILC-SAN under different encoders, $E_{pop\_det}$ achieves the best performance, which is different from the experimental conclusions obtained by previous methods~\cite{zhang2022multiscale}. Considering the significant advantage of energy consumption brought by spike input, we focus on the ILC-SAN using $E_{pop\_det}$, which is the first fully spiking actor network. Meanwhile, its performance outperforms the state-of-the-art methods and the corresponding DANs, which demonstrates that our method has achieved a win-win situation in terms of performance and energy efficiency. Finally, the performance variance of our method is relatively small in most games, which indicates the stability of our method.

The complete learning curves of PopSAN and ILC-SAN using $E_{pop\_det}$ across the first four tasks are shown in Fig. \ref{fig:pop-det}. As we can see, ILC-SAN achieves better performance on Ant-v3 and Hopper-v3. Especially on Hopper-v3, ILC-SAN successfully solves the learning problems encountered by PopSAN. In addition, ILC-SAN achieves the same level of performance on HalfCheetah-v3 and Walker2d-v3.

In Fig. \ref{fig:pop}, we show the complete learning curves of MDC-SAN and ILC-SAN using $E_{pop}$ across the first four tasks. It can be seen that ILC-SAN achieves better performance on Walker2d-v3. In addition, ILC-SAN achieves the same level of performance on Ant-v3, HalfCheetah-v3 and Hopper-v3. Although the performance of ILC-SAN on HalfCheetah-v3 and Hopper-v3 is slightly inferior at the early stage of learning, it still converges to a considerable performance after around 1 million training steps.

\begin{table*}[t]
\caption{Max average rewards over 10 random seeds for various decoding methods without intra-layer connections}
\label{tab:decode}
\centering
\begin{tabular}{cccccc}
\toprule
Decoding Method              & Ant-v3   & HalfCheetah-v3 & Hopper-v3  & Walker2d-v3    & APR \\
\midrule
$D_{fr}$                     & 4848$\pm$1023 & 10523$\pm$1142 & 517$\pm$1057 & 4199$\pm$1426 & 71.94\%\\
\midrule
$D_{last}$ ($\alpha_V=1.0$)  & 5433$\pm$619  & 9932$\pm$1238 & 3249$\pm$1082 & 4668$\pm$702  & \textbf{94.95\%}\\
$D_{last}$ ($\alpha_V=0.75$) & 5001$\pm$667  & 9944$\pm$1215 & 3190$\pm$1075 & 5049$\pm$426  & 94.50\%\\
$D_{last}$ ($\alpha_V=0.5$)  & 5183$\pm$903  & 9936$\pm$1308 & 3217$\pm$1089 & 4910$\pm$654  & 94.81\%\\
\midrule
$D_{max}$ ($\alpha_V=1.0$)   & 5222$\pm$464  & 9540$\pm$1564 & 3506$\pm$158  & 4609$\pm$604  & 94.59\%\\
$D_{max}$ ($\alpha_V=0.75$)  & 5021$\pm$910  & 8775$\pm$1661 & 3453$\pm$156  & 4723$\pm$439  & 92.03\%\\
$D_{max}$ ($\alpha_V=0.5$)   & 5268$\pm$752  & 9694$\pm$1827 & 3200$\pm$1085 & 4843$\pm$797  & 94.16\%\\
\midrule
$D_{mean}$ ($\alpha_V=1.0$)  & 5297$\pm$544  & 9995$\pm$1169 & 2183$\pm$1712 & 4691$\pm$770  & 87.03\%\\
$D_{mean}$ ($\alpha_V=0.75$) & 5635$\pm$350  & 9181$\pm$1301 & 2514$\pm$1410 & 4537$\pm$1388 & 88.21\%\\
$D_{mean}$ ($\alpha_V=0.5$)  & 5278$\pm$809  & 9394$\pm$1351 & 512$\pm$1001  & 4309$\pm$553  & 71.73\%\\
\bottomrule
\end{tabular}
\end{table*}

\begin{table*}[t]
\caption{Max average rewards over 10 random seeds for different decoding methods with intra-layer connections}
\label{tab:dec_intra}
\centering
\begin{tabular}{cccccc}
\toprule
Decoding Method              & Ant-v3       & HalfCheetah-v3  & Hopper-v3  & Walker2d-v3   & APR \\
\midrule
$D_{fr}$                     & 5279$\pm$508 & 10111$\pm$1065 & 1394$\pm$1394 & 4794$\pm$407 & 82.13\%\\
\midrule
$D_{last}$ ($\alpha_V=1.0$)  & 5653$\pm$533  & 10376$\pm$1163 & 3583$\pm$86  & 4511$\pm$462 & \textbf{98.61\%}\\
$D_{last}$ ($\alpha_V=0.75$) & 4796$\pm$751  & 10304$\pm$803  & 3527$\pm$163 & 4563$\pm$835 & 94.38\%\\
$D_{last}$ ($\alpha_V=0.5$)  & 5341$\pm$532  & 10200$\pm$1399 & 3479$\pm$128 & 4955$\pm$664 & 98.24\%\\
\bottomrule
\end{tabular}
\end{table*}

\subsection{Analysis of Decoding Methods}
\label{sec:analysis}

We begin by evaluating the influence of various decoding methods on performance while removing the intra-layer connections in the ILC-SAN. Due to the relationship between the performance of membrane voltage coding methods and the voltage decay factor of non-spiking neurons, we choose three representative values of the voltage decay factor, $\alpha_V=1.0$, $0.75$, $0.5$. As shown in Table \ref{tab:decode}, we compare our membrane voltage coding methods with the mainstream decoding method based on firing rate ($D_{fr}$)~\cite{tang2020reinforcement,zhang2022multiscale} using $E_{pop\_det}$ and demonstrate that both $D_{last}$ and $D_{max}$ are superior to $D_{fr}$ in performance. Compared with $D_{last}$ and $D_{max}$, the performance of $D_{mean}$ is relatively poor. When we deploy the model to the neuromorphic chips, $D_{last}$ only needs to retain the membrane voltage at the last time-step after executing $T$ time-steps. Due to the advantages of deployment and performance, we finally adopt $D_{last}$ as the decoding scheme.

Then we compare the decoding methods ($D_{last}$ and $D_{fr}$) on performance after adding intra-layer connections into ILC-SAN using $E_{pop\_det}$. As Table \ref{tab:dec_intra} shows, $D_{last}$ has significant performance advantages over $D_{fr}$. When $\alpha_V=1.0$, $D_{last}$ achieves the optimal performance, which is taken as the final configuration of our ILC-SAN.

\begin{table*}[t]
\caption{Max average rewards over 10 random seeds for ablation study on intra-layer connections}
\label{tab:intra}
\centering
\begin{tabular}{cccccc}
\toprule
Intra-layer Connections & Ant-v3        & HalfCheetah-v3 & Hopper-v3     & Walker2d-v3   & APR \\
\midrule
No $I_{intra}$          & 5433$\pm$619  & 9932$\pm$1238  & 3249$\pm$1082 & 4668$\pm$702  & 94.95\%\\
$I_{self}$              & 5245$\pm$769  & 10168$\pm$1465 & 3512$\pm$146  & 4843$\pm$757  & 97.40\%\\
$I_{lateral}$           & 5389$\pm$747  & 9384$\pm$987   & 3521$\pm$145  & 4702$\pm$728  & 95.55\%\\
$I_{bias}$              & 4935$\pm$862  & 9906$\pm$1208  & 3141$\pm$1034 & 4409$\pm$1527 & 90.56\%\\
$I_{lateral}+I_{bias}$  & 5549$\pm$404  & 9685$\pm$998   & 3464$\pm$151  & 4851$\pm$907  & 97.34\%\\
$I_{self}+I_{bias}$     & 5114$\pm$729  & 8431$\pm$1410  & 2645$\pm$1439 & 4699$\pm$504  & 85.76\%\\
$I_{self}+I_{lateral}$  & 5067$\pm$727  & 9117$\pm$2923  & 3525$\pm$133  & 4788$\pm$745  & 93.90\%\\
$I_{intra}$             & 5653$\pm$533  & 10376$\pm$1163 & 3583$\pm$86   & 4511$\pm$462  & \textbf{98.61\%}\\
\bottomrule
\end{tabular}
\end{table*}

\subsection{Analysis of Intra-layer Connections}
\label{sec:intra}

In this section, we analyze the impact of intra-layer connections. As shown in Eq. (\ref{eq:intra}), all the elements on the main diagonal of $\bm{W}^m_{intra}$ represent the weight of self connections and other elements represent the weight of lateral connections. To evaluate the contribution of these two parts as well as $\bm{b}_{intra}^m$, we introduce $I_{intra}$, $I_{self}$, $I_{lateral}$ and $I_{bias}$ here. $I_{intra}$ represents our full method which means the current received from the intra-layer connection of any neuron in the output population, as well as the intra-layer bias current. That is, $\bm{I}_{t+1}^{m}$ is formed by $I_{intra}$ of each neuron in the $m$-th output population at time-step $t+1$. $I_{self}$ means only the elements on the main diagonal of $\bm{W}^m_{intra}$ are retained, while $I_{lateral}$ means only the elements that are not on the main diagonal are retained. $I_{bias}$ represents $\bm{W}_{intra}^m\bm{S}_t^{L,m}$ is removed and only $\bm{b}_{intra}^m$ is retained. Overall, $I_{intra}=I_{self}+I_{lateral}+I_{bias}$. We further evaluate each part of $I_{intra}$ on all four tasks while keeping the input coding methods fixed to $E_{pop\_det}$ and the decoding methods fixed to $D_{last}$ with $\alpha_V=1.0$.

As shown in Table \ref{tab:intra}, we conduct the ablation study on intra-layer connections, which proves that each component is indispensable and complementary in performance. Using the complete $I_{intra}$, our model achieves the best performance, which increases the APR by 3.66\% compared with that without $I_{intra}$. This result verifies that the spatial-temporal information brought by intra-layer connections effectively improves the representation capacity of the network and led to better action representation. For individual components, $I_{self}$ can bring the greatest performance improvement as it enriches the dynamics of spiking neurons. However, when $I_{self}$ is combined with either $I_{lateral}$ or $I_{bias}$, the performance of the model degrades greatly. These results show that $I_{lateral}$ and $I_{bias}$ have the strongest complementary effect and cannot be used alone, which is also supported by the experimental results of $I_{lateral}+I_{bias}$. Therefore, the performance improvement brought by intra-layer connections mainly comes from two parts, namely, the current based on the neuron's dynamics $I_{self}$ and the intra-layer current from outside the neuron $I_{lateral}+I_{bias}$.

\begin{table*}[t]
\caption{Max average rewards over 10 random seeds for comparative experiments on the backbone SNN}
\label{tab:layer}
\centering
\begin{tabular}{cccccc}
\toprule
$N_{hidden}$            & Ant-v3        & HalfCheetah-v3 & Hopper-v3     & Walker2d-v3   & APR \\
\midrule
1                       & 4205$\pm$809  & 7455$\pm$549   & 3014$\pm$719  & 4026$\pm$621  & 78.55\%\\
2                       & 5653$\pm$533  & 10376$\pm$1163 & 3583$\pm$86   & 4511$\pm$462  & \textbf{98.61\%}\\
3                       & 5402$\pm$722  & 10580$\pm$1727 & 3192$\pm$951  & 4623$\pm$724  & 95.73\%\\
4                       & 4760$\pm$997  & 10057$\pm$1081 & 3223$\pm$776  & 4221$\pm$1281 & 89.76\%\\
\bottomrule
\end{tabular}
\end{table*}

\begin{table*}[t]
\caption{The average firing rate of the neuronal layer before each fully-connected layer for PopSAN and ILC-SAN}
\label{tab:fr}
\centering
\begin{tabular}{ccccccc}
\toprule
Task                            & Actor Network & FC1    & FC2    & FC3    & Group FC & Intra FC \\
\midrule
\multirow{2}{*}{Ant-v3}         & PopSAN        & 3.0\%  & 38.8\% & 58.3\% & 83.9\%   & $-$      \\
                                & ILC-SAN       & 2.8\%  & 35.5\% & 49.0\% & 41.3\%   & 31.5\%   \\
\midrule
\multirow{2}{*}{HalfCheetah-v3} & PopSAN        & 19.7\% & 33.5\% & 51.0\% & 69.4\%   & $-$      \\
                                & ILC-SAN       & 18.7\% & 25.9\% & 39.1\% & 40.0\%   & 30.7\%   \\
\midrule
\multirow{2}{*}{Hopper-v3}      & PopSAN        & 29.8\% & 68.7\% & 56.7\% & 45.9\%   & $-$      \\
                                & ILC-SAN       & 16.5\% & 20.8\% & 36.9\% & 42.1\%   & 32.8\%   \\
\midrule
\multirow{2}{*}{Walker2d-v3}    & PopSAN        & 18.7\% & 51.2\% & 67.8\% & 52.8\%   & $-$      \\
                                & ILC-SAN       & 15.9\% & 27.4\% & 47.5\% & 47.1\%   & 37.0\%   \\
\bottomrule
\end{tabular}
\end{table*}

\begin{table*}[t]
\caption{The number of operations of each fully-connected layer for PopSAN and ILC-SAN}
\label{tab:op}
\centering
\begin{tabular}{cccccccc}
\toprule
Task                            & Actor Network & FC1       & FC2        & FC3       & Group FC & Intra FC & Total \\
\midrule
\multirow{2}{*}{Ant-v3}         & PopSAN        & 42624 SOP & 127140 SOP & 59699 SOP & 336 FLOP & $-$      & 229463 SOP$+$336 FLOP\\
                                & ILC-SAN       & 40351 SOP & 116326 SOP & 50176 SOP & 165 SOP  & 1260 SOP & 208278 SOP \\
\midrule
\multirow{2}{*}{HalfCheetah-v3} & PopSAN        & 42867 SOP & 109773 SOP & 39168 SOP & 208 FLOP & $-$      & 191808 SOP$+$208 FLOP \\
                                & ILC-SAN       & 40691 SOP & 84869 SOP  & 30029 SOP & 120 SOP  & 921 SOP  & 156630 SOP  \\
\midrule
\multirow{2}{*}{Hopper-v3}      & PopSAN        & 41958 SOP & 225116 SOP & 21773 SOP & 69 FLOP  & $-$      & 288847 SOP$+$69 FLOP \\
                                & ILC-SAN       & 23232 SOP & 68157 SOP  & 14170 SOP & 63 SOP   & 492 SOP  & 106114 SOP \\
\midrule
\multirow{2}{*}{Walker2d-v3}    & PopSAN        & 40604 SOP & 167772 SOP & 52070 SOP & 158 FLOP & $-$      & 260447 SOP$+$158 FLOP  \\
                                & ILC-SAN       & 34598 SOP & 89784 SOP  & 36480 SOP & 141 SOP  & 1110 SOP & 162114 SOP \\
\bottomrule
\end{tabular}
\end{table*}

\begin{table}[t]
\caption{Energy consumption of different tasks per inference for PopSAN and ILC-SAN. The unit of energy is nanojoule (nJ)}
\label{tab:power}
\centering
\begin{tabular}{cccc}
\toprule
Task           & DAN     & PopSAN & ILC-SAN \\
\midrule
Ant-v3         & 1200.0  & 18.7   & 16.0  \\
HalfCheetah-v3 & 892.8   & 15.5   & 12.1  \\
Hopper-v3      & 864.0   & 22.6   & 8.2   \\
Walker2d-v3    & 892.8   & 20.8   & 12.5  \\
\bottomrule
\end{tabular}
\end{table}

\subsection{Analysis of the Backbone SNN}

In previous experiments, we adopt the mainstream setting of TD3 algorithms, which involves two hidden layers with 256 nodes in the backbone SNN~\cite{tang2021deep,zhang2022multiscale}. The hidden layers are located between the input layer and the output layer. To evaluate the effects of network depth, we have conducted an additional experiment on the number of hidden layers in the backbone SNN ($N_{hidden}$). In this experiment, the number of nodes in each hidden layer is set to 256. As shown in Table \ref{tab:layer}, the performance of ILC-SANs reaches its peak when $N_{hidden}=2$. In general, the larger network could obtain higher performance. However, in RL tasks, the larger network could bring overfitting due to the sensitive learning targets of the temporal-difference loss and the unstable training data~\cite{ota2021training,kumar2020implicit}.

\subsection{Analysis of Energy Consumption}

Since low energy consumption is the main advantage of SNNs, we estimate the energy consumption of different networks in this section. Due to the floating-point input, the first fully-connected layer of MDC-SAN cannot be directly deployed to the neuromorphic chips. In addition, due to the high proportion of the computation of the first fully-connected layer in the entire network, considering the simulation time $T$, the energy consumption of MDC-SAN will be even higher than that of the original DAN. Therefore, we only calculate the theoretical energy consumption of DAN, PopSAN and our ILC-SAN. 

Taking the hardware selection in ~\cite{hu2018spiking} as a reference, we employ the FPGA of Intel Stratix 10 TX and the neuromorphic chip of ROLLS~\cite{qiao2015reconfigurable} for estimation. As reported in ~\cite{hu2018spiking}, the FPGA operates at a cost of 12.5pJ per FLOP (floating-point operation) and the ROLLS consumes 77fJ per SOP (synaptic operation~\cite{merolla2014million}). For a fully-connected layer, assuming that the size of the input sample and the output sample is $Dim_{in}$ and $Dim_{out}$, the number of operations required is $Dim_{in}Dim_{out}$ FLOPs for ANN and $Tfr_{in}Dim_{in}Dim_{out}$ SOPs for SNN, where $T$ is the simulation time and $fr_{in}$ is the average firing rate of the previous neuronal layer. Note that we use ($Dim_{in}$, $Dim_{out}$) to represent the fully-connected layer later.

First, we need to determine the number of operations required for each task (FLOP for floating-point input and SOP for spike input). Table \ref{tab:task} shows the state dimension and action dimension for different tasks, which affects the network structure and the number of operations. For DAN, the number of operations is equal to $256(256+N+M)$, which goes as follows: 96000 FLOP, 71424 FLOP, 69120 FLOP, 71424 FLOP respectively for Ant-v3, HalfCheetah-v3, Hopper-v3 and Walker2d-v3. For PopSAN, the first three fully-connected layers can be expressed as FC1 ($N\cdot P_{in}$, 256), FC2 (256, 256), FC3 (256, $M\cdot P_{out}$), and the population decoder, which consists of a group of $M$ fully-connected layers ($P_{out}$, 1), can be expressed as Group FC. For our ILC-SAN, we have an additional calculation of intra-layer connections, which consists of a group of $M$ fully-connected layers ($P_{out}$, $P_{out}$) and can be expressed as Intra FC. As shown in Fig. \ref{fig:san}, the learnable layer represents the Group FC, and the intra-layer connections represent the Intra FC. As shown in Table \ref{tab:fr}, we run all the trained models for one episode to obtain the average firing rate of each neuronal layer. Although the neuronal layer before Group FC is the same as that before Intra FC, the average firing rate is different due to the time-steps involved in the calculation. It can be seen that compared with PopSAN, the firing rate of spiking neurons in each layer of ILC-SAN is significantly reduced. 

Then we estimate the number of operations for each fully-connected layer (Table \ref{tab:op}), based on the formula of SOP for the fully-connected layer mentioned in this section. Since the input of Group FC in PopSAN is the floating-point firing rate, it needs to be calculated on traditional hardware, using FLOP to count the number of operations.

Concerning each platform’s energy efficiency, we calculate the energy consumption required by DAN, PopSAN and ILC-SAN per inference. As Table \ref{tab:power} shows, the energy consumption of DAN is more than 71 times greater than the energy consumed by ILC-SAN across all four tasks. In addition, compared with PopSAN, ILC-SAN can reduce 35.1\% of the energy consumption per inference on average.

\section{Conclusion}

In this paper, we present ILC-SAN, a fully spiking actor network with intra-layer connections for the reinforcement learning. With the help of membrane voltage coding, we solve the deployment problem of current spike-based RL methods, and provide a better alternative for the representation of floating-point values, such as continuous action and Q-value. In addition, we add intra-layer connections to the output population, and analyze the effects of self connections and lateral connections, both of which have improved the performance of the model. In terms of energy efficiency, our model effectively reduces the firing rate of spiking neurons for each layer, thus greatly reducing the energy consumption running on neuromorphic hardware. Our model is evaluated on eight continuous control tasks from OpenAI gym, and outperforms the start-of-the-art methods. We provide a comparison of our method with the corresponding deep network in terms of performance, stability, and energy consumption, comprehensively demonstrating the superiority of our method.

As an energy-efficient alternative for realtime robot control tasks, our ILC-SAN has great application potential, which could be taken as the foundation for the follow-up work. For example, one of the future directions is to combine ILC-SAN with neuromorphic sensors to achieve robot control in real-world scenes, which requires addressing the limitations of vector input in our method and extending it to higher-order tensor inputs such as images or spatiotemporal signals. In the future, more biologically plausible rules could be applied to SNNs to improve performance and energy efficiency for different tasks. We believe that neuroscience will be an important source of inspiration for artificial intelligence.

\bibliographystyle{IEEEtran}
\bibliography{main}

\end{document}